\documentclass[letterpaper]{article} 
\usepackage{aaai2026}
\usepackage{times}  
\usepackage{helvet}  
\usepackage{courier}  
\usepackage[hyphens]{url}  
\usepackage{graphicx} 
\usepackage{natbib}  
\usepackage{caption} 
\usepackage{booktabs}
\usepackage{multirow}
\usepackage{algorithm}
\usepackage{algpseudocode}
\usepackage{graphicx}
\usepackage{booktabs}
\usepackage{amssymb}
\usepackage{amsmath}
\usepackage{amsfonts}
\usepackage{pifont}
\definecolor{bestcell}{gray}{0.90} 
\usepackage{newfloat}
\usepackage{listings}

\DeclareCaptionStyle{ruled}{labelfont=normalfont,labelsep=colon,strut=off} 
\floatstyle{ruled}
\newfloat{listing}{tb}{lst}{}
\floatname{listing}{Listing}
\definecolor{confICLR}{HTML}{FFFFFF} 
\definecolor{confNeurIPS}{HTML}{FFFFFF} 
\definecolor{journalNC}{HTML}{FFFFFF} 
\definecolor{confARXIV}{HTML}{FFFFFF} 
\definecolor{confMICCAI}{HTML}{FFFFFF} 
\definecolor{confCVPR}{HTML}{FFFFFF} 
\definecolor{confICCV}{HTML}{FFFFFF} 
\definecolor{confACMMM}{HTML}{FFFFFF} 

\newcommand{\ConfTag}[3]{\colorbox{#1}{\textcolor{black}{\tiny\textsc{#2 #3}}}}

\newcommand{\ICLR}[1]{\ConfTag{confICLR}{ICLR}{#1}}
\newcommand{\NC}[1]{\ConfTag{journalNC}{Nature Communications}{#1}}
\newcommand{\ARXIV}[1]{\ConfTag{confARXIV}{arXiv}{#1}}

\def\method{\textsc{NeuralOM}} 

\title{\method{}: Neural Ocean Model for Subseasonal-to-Seasonal Simulation}

\author{
Yuan Gao\textsuperscript{\rm 1}\equalcontrib,
Hao Wu\textsuperscript{\rm 1}\equalcontrib\thanks{Project lead and technical guidance.},
Fan Xu\textsuperscript{\rm 2},
Yanfei Xiang\textsuperscript{\rm 1},
Ruijian Gou\textsuperscript{\rm 3},\\
Ruiqi Shu\textsuperscript{\rm 1},
Qingsong Wen\textsuperscript{\rm 4},
Xian Wu\textsuperscript{\rm 5}\footnotemark[3],
Kun Wang\textsuperscript{\rm 6}\footnotemark[3],
Xiaomeng Huang\textsuperscript{\rm 1}\thanks{Corresponding author.}
}
\affiliations{
\textsuperscript{\rm 1}Tsinghua University,\\
\textsuperscript{\rm 2}Shenzhen Loop Area Institute,\\
\textsuperscript{\rm 3}Ocean University of China,\\
\textsuperscript{\rm 4}Squirrel Ai Learning,\\
\textsuperscript{\rm 5}Tencent,\\
\textsuperscript{\rm 6}Nanyang Technological University

\{yuangao24, wu-h25, xiangyf22, srq24\}@mails.tsinghua.edu.cn; fanxu@slai.edu.cn; grj@stu.ouc.edu.cn; qingsongwen@squirrelai.com; kevinxwu@tencent.com; wang.kun@ntu.edu.sg; hxm@tsinghua.edu.cn  

}

\usepackage{bibentry}

\begin{document}
\maketitle
\begin{abstract}
Long-term, high-fidelity simulation of slow-changing physical systems, such as the ocean and climate, presents a fundamental challenge in scientific computing. Traditional autoregressive machine learning models often fail in these tasks as minor errors accumulate and lead to rapid forecast degradation. To address this problem, we propose \textbf{\method{}}, a general neural operator framework designed for simulating complex, slow-changing dynamics. \method{}'s core consists of two key innovations: (1) a \textbf{\textit{Progressive Residual Correction Framework}} that decomposes the forecasting task into a series of fine-grained refinement steps, effectively suppressing long-term error accumulation; and (2) a \textbf{\textit{Physics-Guided Graph Network}} whose built-in \textbf{adaptive messaging mechanism} explicitly models multi-scale physical interactions, such as gradient-driven flows and multiplicative couplings, thereby enhancing physical consistency while maintaining computational efficiency. We validate \method{} on the challenging task of global Subseasonal-to-Seasonal (S2S) ocean simulation. Extensive experiments demonstrate that \method{} not only surpasses state-of-the-art models in forecast accuracy and long-term stability, but also excels in simulating extreme events. For instance, at a 60-day lead time, \method{} achieves a 13.3\% lower RMSE compared to the best-performing baseline, offering a stable, efficient, and physically-aware paradigm for data-driven scientific computing.
\end{abstract}

\vspace{-10pt}
\begin{links}
\link{Code}{https://github.com/YuanGao-YG/NeuralOM}
\end{links}
\vspace{-15pt}

\section{Introduction}

\begin{figure}[!t]
\centering
\includegraphics[width=0.47\textwidth]{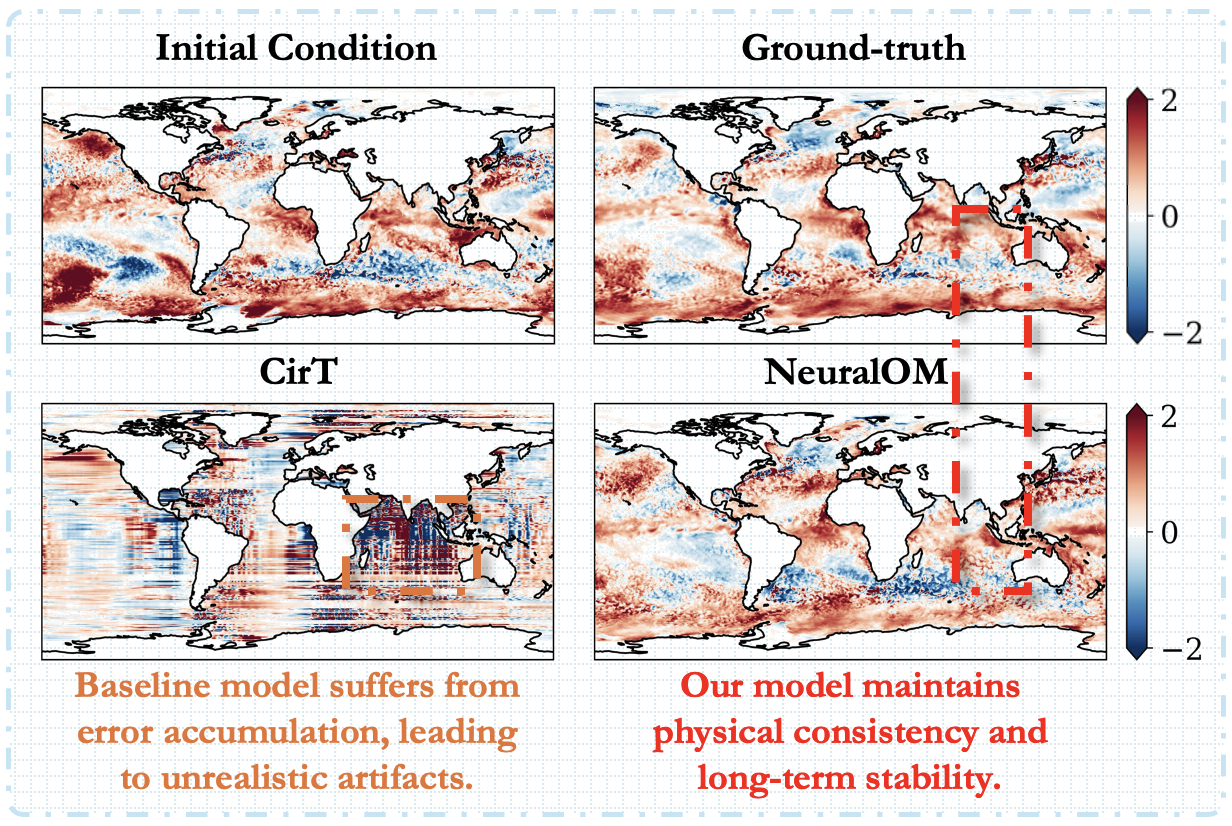}
\vspace{-15pt}
  \caption{
    NeuralOM overcomes the long-term instability challenge in simulating slow-changing physical systems.
    Shown is a 60-day Subseasonal-to-Seasonal (S2S) simulation for sea surface temperature anomaly. 
    \textbf{(\textit{Bottom Left})} A state-of-the-art baseline (CirT) collapses due to compounding errors, producing physically unrealistic artifacts and failing to capture the ocean state.
    \textbf{(\textit{Bottom Right})} In contrast, NeuralOM maintains high fidelity and long-term stability, accurately reproducing the complex patterns of the ground-truth, as highlighted in the detailed view (red dashed lines).
  }
  \vspace{-20pt}
\label{fig:intro_1}
\end{figure}

The long-term simulation of complex physical systems~\cite{wang2024modeling, wu2025advanced}, such as the global ocean and climate, is a cornerstone of modern science yet remains a great challenge.
While machine learning (ML) models offer a computationally efficient path to complement traditional numerical solvers, they face a fundamental bottleneck in such tasks: \textbf{\textit{long-term stability}}.
This issue is especially severe in \textbf{\textit{slow-changing systems}}, where the state evolves in a subtle manner.
In standard autoregressive frameworks, minor prediction errors at each step inevitably accumulate, causing the simulation to diverge catastrophically from the true physical dynamics, a well-known problem of \textbf{compounding error}.
This error cascade often leads data-driven models to fail in long-term rollouts. As shown in Figure~\ref{fig:intro_1} (bottom left), a state-of-the-art baseline model generates substantial, physically unrealistic artifacts in its long-range forecast.

To address this challenge, existing machine learning approaches have been explored, but they are limited in two key aspects.
\ding{182}. \textit{\textbf{First}}, they often struggle to capture the fine-grained dynamics inherent to slow-changing systems.
Many models are designed to predict sharp, significant shifts, making them ill-suited for learning the small, incremental updates that define ocean dynamics.
This typically results in either over-smoothing the predictions or failing to capture critical physical anomalies.
\ding{183}. \textit{\textbf{Second}}, the internal mechanisms of these models are often physics-agnostic.
For instance, standard message passing in Graph Neural Networks (GNNs)~\cite{keisler2022forecasting} treats all node interactions uniformly, unable to distinguish between fundamentally different physical processes like gradient-driven flows and multiplicative couplings between variables.
Furthermore, they lack the ability to adaptively model phenomena across different spatio-temporal scales, from large-scale, slow-moving currents to small-scale, energetic eddies.

To systematically address these issues, we introduce \textbf{\method{}}, a neural operator framework designed for stable, high-fidelity simulation of slow-changing dynamics. 
\method{}'s core is built on two key innovations.
First, we propose a \textbf{Progressive Residual Correction Framework}. This framework decomposes the complex forecasting task into a cascade of refinement steps, where each step focuses on learning and correcting the residual error from the previous one. 
This strategy of ``small, iterative corrections" enables the model to capture extremely fine-grained dynamic changes, thereby effectively suppressing long-term error accumulation.
Second, we design a novel \textbf{Physics-Guided Graph Network} with a built-in \textbf{adaptive messaging mechanism}.
It incorporates physical priors into the network by explicitly modeling gradient-driven flows and multiplicative couplings between variables. 
Crucially, its aggregation scheme is not fixed but dynamically adjusts, allowing it to inherently handle multi-scale dynamics.

We validate \method{} on the challenging benchmark of global Subseasonal-to-Seasonal (S2S) ocean simulation and forecasting. Extensive experiments provide strong evidence of its superior performance:

\noindent \textbf{Experimental Observations}. \ding{182} \textbf{\textit{State-of-the-art accuracy in long-term simulation}:} As shown in Table~\ref{tab:mainres}, in $60$-day simulation, \method{} achieves a normalized RMSE of just $0.7014$, a \textbf{$13.3$\%} reduction compared to the best-performing baseline, WenHai ($0.8091$). It also improves the Anomaly Correlation Coefficient (ACC) by \textbf{$12.0$\%}, indicating more accurate tracking of the physical field's evolution. \ding{183} \textbf{\textit{Superior visual fidelity and physical consistency}:} As illustrated in Figure~\ref{fig:intro_1} and Figure~\ref{fig:visual}, unlike baseline models like CirT that collapse due to error explosion, \method{} produces physically consistent forecasts that remain highly aligned with the ground truth even at $60$ days, successfully preserving critical fine-scale structures. \ding{184} \textbf{\textit{Validated effectiveness of our designs through ablation studies}:} As detailed in Table~\ref{tab:ablation_mdoel} and Table~\ref{tab:ablation_var}, removing any of our core components, either the Progressive Residual Correction Framework or the Physics-Guided Adaptive Graph Messaging, leads to a significant drop in performance. In particular, incorporating climatology priors improves the ACC metric by more than \textbf{$5$-fold}, highlighting that every design choice is crucial to achieve accurate and stable simulations.

In summary, our main contributions are threefold: (1) a progressive residual correction framework designed for slow-changing systems; (2) a physics-guided graph network with an adaptive, multi-scale messaging mechanism; and (3) state-of-the-art performance on the S2S ocean simulation benchmark, establishing a new paradigm for data-driven scientific computing.
\section{Related Work}

In recent years, previous works in spatio-temporal mining have achieved stupendous achievements~\cite{pfaff2020learning, li2020fourier, wu2023earthfarseer, wu2024pastnet, wang2024nuwadynamics, wu2024prometheus, wu2024pure, wu2025turb, wu2025spatiotemporal, wu2025differential, jia2025learning}. However, many of these methods are only applicable in idealized or relative simple scenario, whereas ocean system is a complex dynamical system in the real world, requiring the design of specialized methods for accurate prediction~\cite{wu2024neural}. Research in data-driven ocean simulation is primarily divided into two streams: task-specific predictive models and general-purpose global foundation models.

\paragraph{Task-Specific ML Models.} 
These models aim to simulate specific oceanic or atmospheric phenomena and have achieved impressive results in their respective domains. 
Examples include work on El Nio-Southern Oscillation (ENSO) prediction~\cite{ham2019deep, hu2021deep, chang2023predicting, lyu2024resonet, chen2025combined}, as well as models for the Madden-Julian Oscillation (MJO)~\cite{kim2021deep, delaunay2022interpretable, yang2023exploring, shin2024data, shin2024deep} and Marine Heatwaves (MHWs)~\cite{jacox2022global, sun2024deep, parasyris2025marine}. 
The success of these models validates the immense potential of machine learning for complex geoscience problems. 
However, their limitation is also apparent: they are typically highly optimized for a single objective and often overlook the complex interactions between different physical variables and across multiple scales. 
This makes them difficult to generalize into a foundation model capable of capturing the full dynamics of the ocean system.

\begin{figure*}    
\centering
\includegraphics[width=1\linewidth]{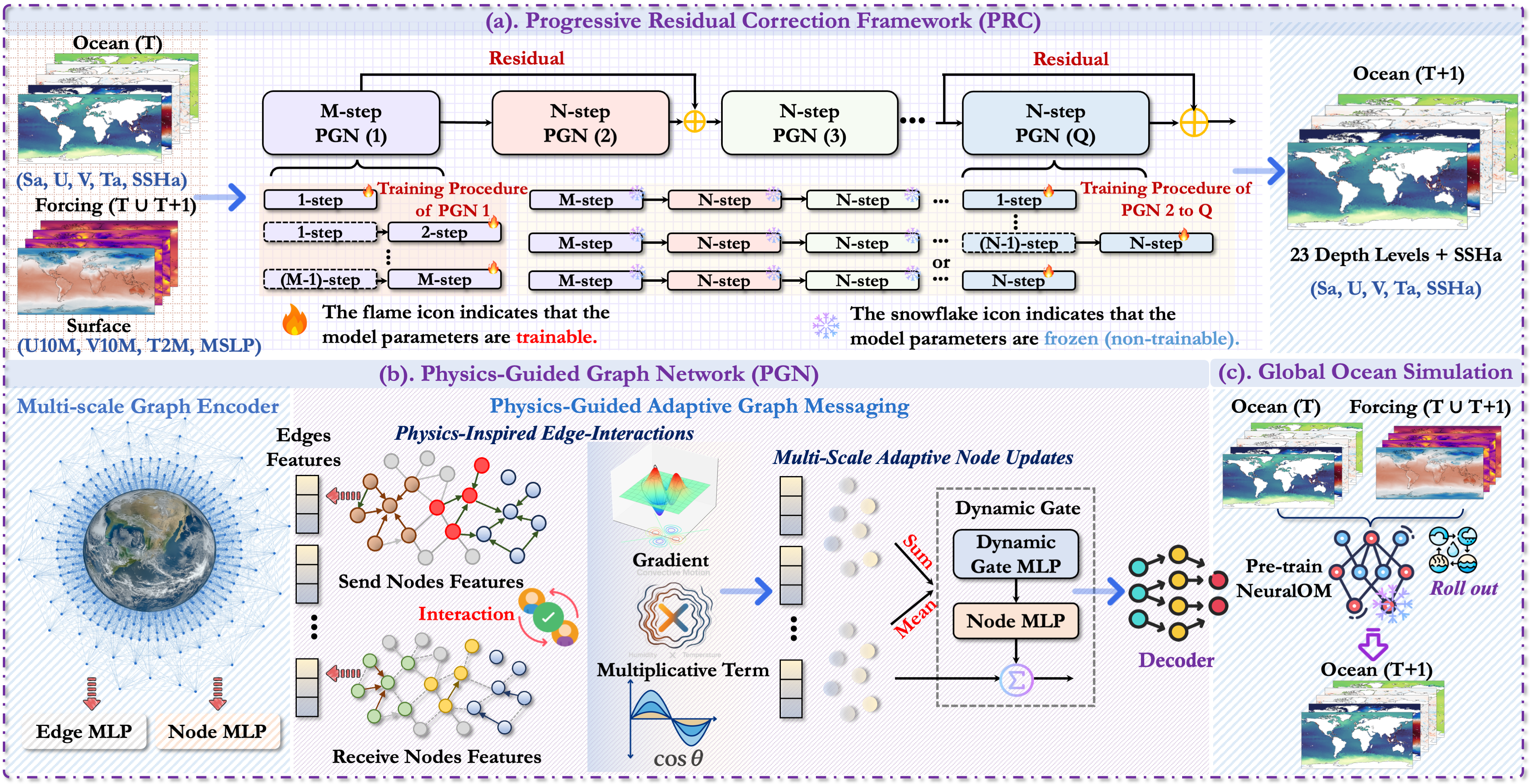}
\vspace{-15pt}
\caption{Overview of Our~\method{}. \textbf{(a)} The overall architecture of the progressive residual correction framework, input variables (subtracting climatology for periodic variables), progressive residual correction stage, and visualization of simulation ocean variables; \textbf{(b)} The proposed physics-guided graph network; \textbf{(c)} The global ocean simulation module uses a rollout approach to generate future results.}
\label{fig_NeuralOM}
\vspace{-15pt}
\end{figure*}

\paragraph{Global Ocean System Foundation Models.} 
Inspired by the success of foundation models in atmospheric science~\cite{pathak2022FourCastNet, bi2023accurate, chen2023fengwu, chen2023fuxi, lam2023learning, oskarsson2024probabilistic, nguyen2024scaling, kochkov2024neural, gao2025oneforecast}, a series of global ocean foundation models have recently emerged, such as AI-GOMS~\cite{xiong2023ai}, Xihe~\cite{wang2024xihe}, LangYa~\cite{yang2024langya}, and WenHai~\cite{cui2025forecasting}. 
These models aim to build a unified digital twin of the ocean and have already surpassed traditional numerical methods in computational efficiency and, in some cases, accuracy~\cite{hao2025deep}.
However, they still face the two core challenges we identified in the introduction. 
{First,} they generally adopt standard autoregressive frameworks and have not fundamentally solved the problem of {error accumulation} in long-term simulations, leading to a decline in forecast fidelity over time.
{Second,} their designs often lack sufficient {physical constraints}. 
For example, while models like WenHai~\cite{cui2025forecasting} incorporate domain knowledge into a Swin Transformer~\cite{liu2021swin}, their feature interaction mechanisms remain relatively {physics-agnostic}, failing to explicitly model physical processes like gradient changes or variable couplings, nor can they adaptively handle multi-scale dynamics.

\method{} is designed precisely to address these limitations. Unlike task-specific models, it is a general-purpose global model. Compared to existing global foundation models, \method{} introduces two fundamental breakthroughs: (1) Its \textbf{Progressive Residual Correction Framework} is a novel architecture specifically designed to suppress error accumulation in slow-changing systems and ensure long-term stability. (2) Its \textbf{Physics-Guided Graph Network}, with its adaptive messaging mechanism, explicitly integrates physical interactions (e.g., gradients and couplings) into the model, addressing the physics-agnostic problem of existing models.

\section{Methodology}

\paragraph{Preliminaries.} We formulate the ocean system simulation as an autoregressive forecasting problem. Given the system state $\mathbf{O}_t$ at the current time step $t$ (such as ocean temperature and salinity) and the external forcing $\mathbf{F}_t$, $\mathbf{F}_{t+1}$ (e.g., atmospheric fields), our goal is to predict the system state at the next time step, $\hat{\mathbf{O}}_{t+1}$. This process is represented by a neural network $\mathcal{M}$:
\begin{equation}
    \hat{\mathbf{O}}_{t+1} = \mathcal{M}(\mathbf{O}_t, \mathbf{F}_t, \mathbf{F}_{t+1}; \Theta),
\end{equation}
where $\Theta$ represents the learnable parameters of the model. During inference, the model performs long-range forecasting by recursively feeding its own prediction from the previous step, $\hat{\mathbf{O}}_{t}$, as input for the next step, starting from an initial state $\mathbf{O}_0$. Further details on the specific variables and data preprocessing used in our experiments are provided in the Appendix.

\subsection{Progressive Correction for Stability}
\label{sec:method_main}

To overcome the challenges of error accumulation and the physics-agnostic nature of existing models, we propose \method{}, a stable and efficient simulation framework designed for slow-changing physical systems. Its core design comprises a \textbf{\textit{Progressive Residual Correction Framework}} and a \textbf{\textit{Physics-Guided Graph Network}} that serves as its core engine.

\subsubsection{Progressive Residual Correction}

When simulating slow-changing systems, directly predicting the entire next state $\mathbf{O}_{t+1}$ is difficult, as the true change ($\mathbf{O}_{t+1} - \mathbf{O}_{t}$) is remarkably small. Most models struggle to capture this subtle signal accurately, leading to rapid error accumulation.

To address this, we design a \textbf{Progressive Residual Correction} framework. It decomposes the complex single-step prediction task into a series of simpler, cascaded correction steps. First, a base model makes an initial, coarse prediction of the next state. Subsequently, a series of residual models are invoked, with each model focusing on learning and correcting the prediction error from the previous stage. This process can be abstractly formulated as:
\begin{align}\small
    \hat{\mathbf{O}}_{t+1}^{(1)} &= \mathcal{M}_{\text{base}}(\mathbf{X}_t), \label{eq:base_pred_revised} \\
    \hat{\mathbf{O}}_{t+1}^{(q)} &= \hat{\mathbf{O}}_{t+1}^{(q-1)} + \mathcal{M}_{\text{residual}}^{(q)}(\hat{\mathbf{O}}_{t+1}^{(q-1)}), \quad \text{for } q=2, \dots, Q, \label{eq:residual_corr_revised}
\end{align}
where $\mathbf{X}_t$ is the input of model, $\hat{\mathbf{O}}_{t+1}^{(1)}$ is the initial prediction, and $\hat{\mathbf{O}}_{t+1}^{(q)}$ is the refined prediction after $q-1$ correction stages. The final output is $\hat{\mathbf{O}}_{t+1} = \hat{\mathbf{O}}_{t+1}^{(Q)}$. This approach shifts the model's focus from predicting a large state vector to capturing fine-grained, physically meaningful anomalies, thereby significantly enhancing long-term stability and fidelity. Additionally, for variables with strong periodicity (e.g., sea salinity, sea temperature, and sea surface height), we first subtract their climatological mean before feeding them into the model to further help it focus on learning small anomaly changes.

\subsubsection{Physics-Guided Adaptive Graph Messaging.}

Each model within this framework ($\mathcal{M}_{\text{base}}, \mathcal{M}_{\text{residual}}$) is realized by our novel \textbf{Physics-Guided Graph Network}. Unlike standard GNNs, our model deeply integrates physical priors into the graph's message passing and node update procedures through its core \textbf{\textit{Physics-guided Adaptive Graph Messaging}} module.

\paragraph{Physics-Inspired Edge-Interactions.} Traditional GNNs are often \textit{physics-agnostic}, typically just concatenating adjacent node features and passing them through an MLP. To more realistically emulate the dynamic relationships between physical variables, we design physics-inspired edge interactions (PEI). The message between two nodes (i.e., the edge feature) is constructed not blindly, but through a set of explicit physical operators, including: (1) \textbf{feature differencing} to model \textbf{gradient-driven flows}; (2) \textbf{multiplicative coupling} to capture the joint effects of variables (like temperature and salinity on density); and (3) \textbf{cosine similarity} to identify similar water masses. These operators capture core relationships in ocean dynamics and provide a strong inductive bias. This process can be abstractly represented as:
\begin{equation}\small
\label{eq:mim}
\mathbf{e}_{uv} = f_{\text{PEI}}(h_u, h_v) = \phi([\underbrace{h_u - h_v}_{\text{Gradient}}, \underbrace{h_u \odot h_v}_{\text{Coupling}}, \underbrace{\text{sim}(h_u, h_v)}_{\text{Similarity}}]),
\end{equation}
where $h_u, h_v$ are node features and $\phi$ is a neural network that fuses these physical features.

\paragraph{Multi-scale Adaptive Node Aggregation.} Physical systems often contain motions at different scales. To capture both large-scale, slow-changing processes and small-scale, energetic events simultaneously, we introduce a \textbf{dynamic gating mechanism} for aggregating neighbor information. This mechanism adaptively arbitrates between \textbf{sum} and \textbf{mean} aggregation based on local dynamics. For localized, high-energy events (like eddies), the gate favors sum aggregation to preserve their intensity. For large-scale, smooth flows, it favors mean aggregation to obtain a more robust, macroscopic representation.
\begin{equation}\small
\label{eq:gate}
h'_v = (1 - \gamma) \cdot \text{Aggregate}_{\text{mean}}(\{\mathbf{e}_{uv}\}) + \gamma \cdot \text{Aggregate}_{\text{sum}}(\{\mathbf{e}_{uv}\}),
\end{equation}
where the gating coefficient $\gamma$ is dynamically learned by the network. This adaptive aggregation allows \method{} to efficiently handle complex multi-scale dynamics within a unified framework. For details on graph construction, encoder, and decoder, please refer to the Appendix.

\subsection{Optimization and Inference}

\paragraph{Objective.} We train each component of \method{} by minimizing the relative L2-norm between the predicted and ground-truth states. The loss function $\mathcal{L}$ is defined as:
\begin{equation}
\mathcal{L}(\Theta)=\mathbb{E}_{t, k}\left[\frac{\left\|\hat{O}_{i, j, k}^{t+1}-O_{i, j, k}^{t+1}\right\|_2}{\left\|O_{i, j, k}^{t+1}\right\|_2}\right],
\end{equation}
where $\hat{O}$ and $O$ are the predicted and ground-truth values, respectively, indexed over time $t$, spatial locations $(i, j)$, and variable channels $k$. $\|\cdot\|_2$ denotes the Euclidean norm over the spatial grid (i, j) for each channel $k$. ${E}_{t, k}$ denotes the empirical expectation taken over training time steps $t$ and variable channels $k$, i.e., averaging the channel-wise relative errors over training samples.

\paragraph{Inference Strategy.} For long-range simulation and forecasting, we employ a standard autoregressive (or ``rollout") strategy. Starting from an initial state $\mathbf{X}_0$, the model continuously feeds its own previous prediction as input for the next step to generate a continuous trajectory spanning several weeks or months. 
For idealized \textit{simulation} experiments, we use ground-truth atmospheric data as forcing. This approach eliminates interference from atmospheric forecast errors, allowing for a fairer assessment of the ocean models' intrinsic performance. 
For more realistic \textit{forecasting} experiments, we drive \method{} with forecast fields generated by a state-of-the-art atmospheric forecasting model, OneForecast~\cite{gao2025oneforecast}.

\begin{algorithm}[H]
\caption{NeuralOM Forward Pass}
\label{alg:neuralom_forward_pass_en}
\begin{algorithmic}[1]
\Require Initial state $\mathbf{O}_0$, forcing sequence $\{\mathbf{F}_t\}_{t=0}^{T}$, total steps $T+1$
\Ensure Predicted trajectory $\{\hat{\mathbf{O}}_t\}_{t=1}^{T}$

\State $\mathbf{O}^{\text{prev}} \leftarrow \mathbf{O}_0$
\For{$t = 0, 1, \dots, T-1$}
    \State $\mathbf{X}_t \leftarrow \text{Concat}(\mathbf{O}^{\text{prev}}, \mathbf{F}_t, \mathbf{F}_{t+1})$ \Comment{Combine current inputs}
    
    \Statex \quad \# Stage 1: Progressive Residual Correction Framework
    \State $\hat{\mathbf{O}}_{t+1}^{(1)} \leftarrow \text{PGN}_{\text{base}}(\mathbf{X}_t)$ \Comment{Initial prediction by the base model}
    \State $\hat{\mathbf{O}}_{\text{intermediate}} \leftarrow \hat{\mathbf{O}}_{t+1}^{(1)}$
    
    \For{$q = 2, \dots, Q$} \Comment{Perform Q-1 rounds of residual correction}
        \State $\text{Residual} \leftarrow \text{PGN}_{\text{residual}}^{(q)}(\hat{\mathbf{O}}_{\text{intermediate}})$ \Comment{Residual model predicts the correction}
        \State $\hat{\mathbf{O}}_{\text{intermediate}} \leftarrow \hat{\mathbf{O}}_{\text{intermediate}} + \text{Residual}$ \Comment{Apply the residual correction}
    \EndFor
    
    \State $\hat{\mathbf{O}}_{t+1} \leftarrow \hat{\mathbf{O}}_{\text{intermediate}}$ \Comment{Get the final single-step prediction}
    
    \State $\mathbf{O}^{\text{prev}} \leftarrow \hat{\mathbf{O}}_{t+1}$ \Comment{Update state for the next autoregressive step}
\EndFor

\State \textbf{return} $\{\hat{\mathbf{O}}_t\}_{t=1}^{T}$
\end{algorithmic}
\end{algorithm}

\begin{table*}[t]
    \vskip 0.15in
    \centering
    \begin{sc}
        \renewcommand{\multirowsetup}{\centering}
        \setlength{\tabcolsep}{1.8pt}
        \begin{tabular}{l|cc|cc|cc|cc}
            \toprule
            \multirow{2}{*}{Model} & \multicolumn{8}{c}{Simulate Horizon} \\
            \cmidrule(lr){2-9}
            & \multicolumn{2}{c}{40-day} & \multicolumn{2}{c}{45-day} & \multicolumn{2}{c}{50-day} & \multicolumn{2}{c}{60-day} \\
            \cmidrule(lr){2-9}
            & RMSE  & ACC & RMSE & ACC & RMSE & ACC & RMSE & ACC \\
            \midrule
            FourCastNet~\cite{pathak2022FourCastNet} {\scriptsize\ARXIV{2022}}
              & 5.6295 & 0.0782
              & $>10$ & 0.0190
              & $>10$ & 0.0046
              & $>10$ & 0.0007 \\
            CirT~\cite{liu2025cirt} {\scriptsize\ICLR{2025}}
              & 2.7510 & 0.0143
              & 2.8686 & 0.0115
              & 2.9514 & 0.0099
              & 3.0631 & 0.0081 \\
            \midrule
            WenHai~\cite{cui2025forecasting} {\scriptsize\NC{2025}}
              & \underline{0.7199} & \underline{0.5065}
              & \underline{0.7467} & \underline{0.4703}
              & \underline{0.7700} & \underline{0.4385}
              & \underline{0.8091} & \underline{0.3851} \\
            \midrule
            \method{} (ours)
              & \textbf{0.6509} & \textbf{0.5297}
              & \textbf{0.6665} & \textbf{0.5009}
              & \textbf{0.6795} & \textbf{0.4754}
              & \textbf{0.7014} & \textbf{0.4314} \\
            \midrule
            \method{} Improvement
              & 9.59\% & 4.58\%
              & 10.74\% & 6.49\%
              & 11.74\% & 8.42\%
              & 13.32\% & 12.01\% \\
            \bottomrule
        \end{tabular}
    \end{sc}
    \vspace{-5pt}
     \caption{Performance comparison of \method{} against baselines on global ocean simulation task. Average RMSE (normalized, $\downarrow$) and ACC ($\uparrow$) across all 93 ocean variables are shown. Best results in \textbf{bold}, second best \underline{underlined}.}
     \label{tab:mainres}
     \vspace{-5pt}

\end{table*}

\begin{figure*}[h]
\centering
\includegraphics[width=1\linewidth]{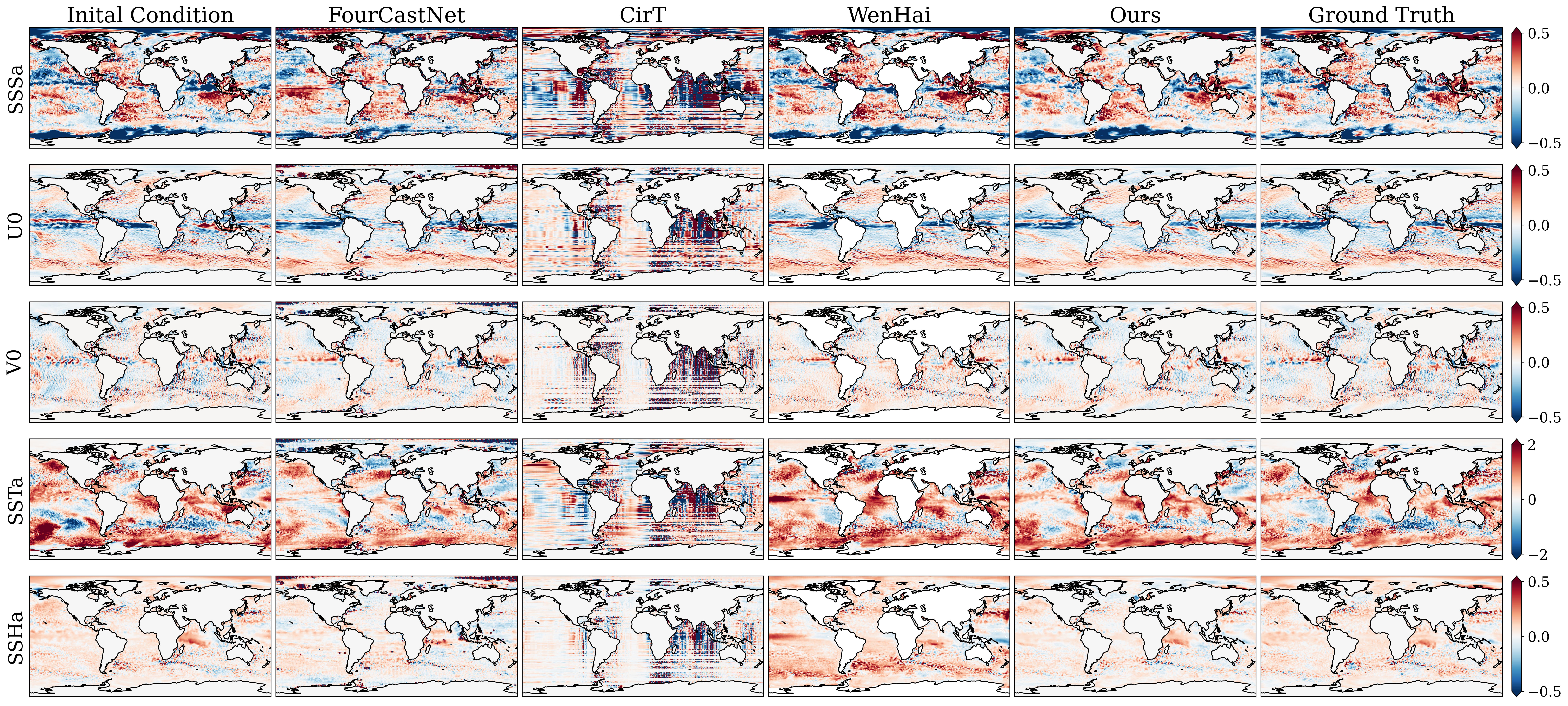}
\vspace{-15pt}
\caption{60-day simulation results of different models.}
\label{fig:visual}
\vspace{-10pt}
\end{figure*}

\paragraph{Summary.} In conclusion, the proposed \method{} framework forms an end-to-end solution for slow-changing physical systems, bridging macroscopic correction strategies with microscopic physical modeling. 
To transparently illustrate its computational flow, we outline the complete forward pass of \method{} in Algorithm~\ref{alg:neuralom_forward_pass_en}. 
In the following section, we will show its practical effectiveness through a series of rigorous benchmarks and ablation studies.

\section{Experiments}

To comprehensively evaluate the performance of \method{} and validate its effectiveness in addressing the challenges of long-term stability and physical consistency, we design a series of rigorous experiments. We first benchmark \method{} against several state-of-the-art models on an idealized global ocean \textbf{simulation} task. We then apply it to a more challenging real-world \textbf{forecasting} scenario. Finally, we dissect the contributions of our model's innovative components through extensive \textbf{ablation studies}. All experiments were conducted on 64 NVIDIA A100 GPUs.

\begin{figure*}[h]
        \centering
        \includegraphics[width=1\linewidth]{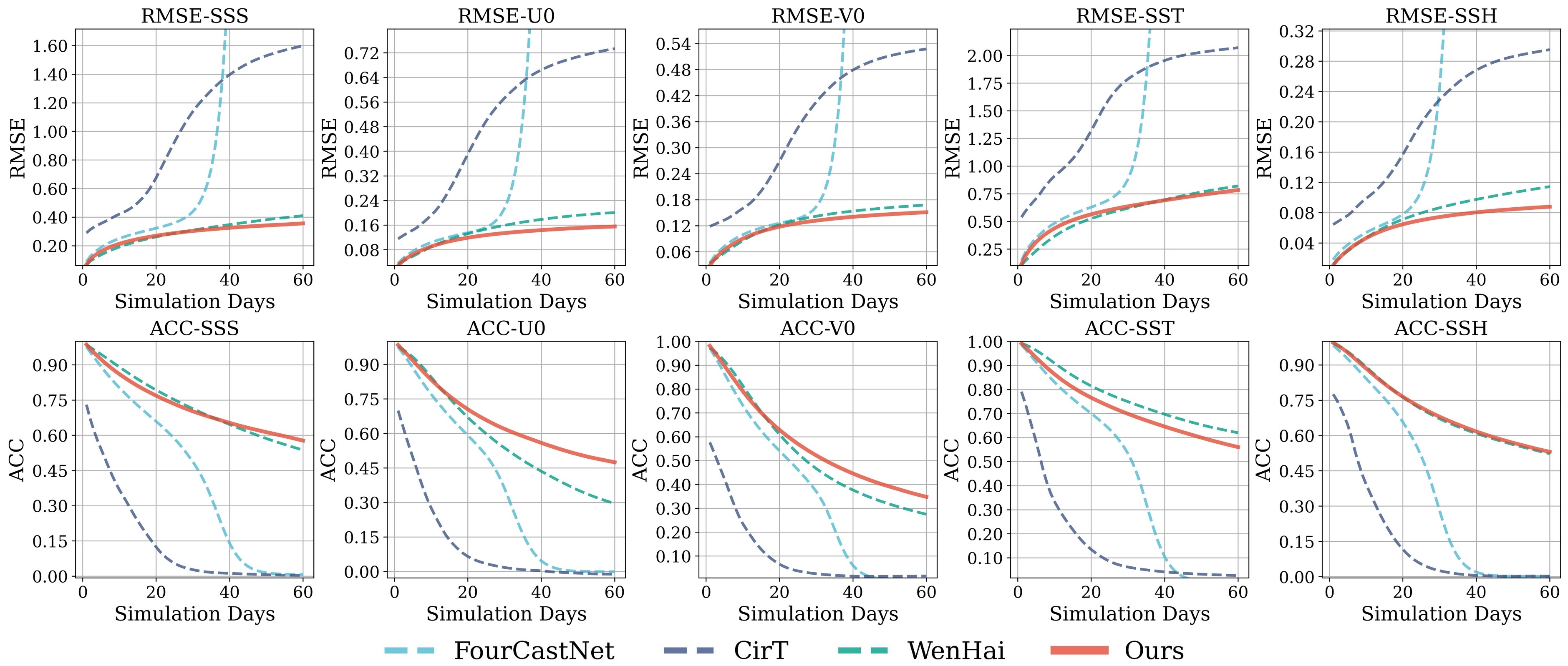}
        \vspace{-20pt}
        \caption{The latitude-weighted RMSE and ACC results of several important ocean surface variables.}
        \label{fig:rmse_acc}
\vspace{-15pt}
\end{figure*}

\subsection{Experimental Setup}

\paragraph{Dataset and Task.} Our experiments are based on the \textbf{GLORYS12} global ocean reanalysis dataset, which provides daily mean ocean states since 1993. We split the data into a training set (1993-2017), a validation set (2018-2019), and a test set (2020). The core task is autoregressive forecasting at the \textbf{Subseasonal-to-Seasonal (S2S)} scale, with rollout horizons ranging from several days up to 60 days. The input consists of 93 ocean state variables (including salinity, velocity, temperature at various depths, and sea surface height) and 8 atmospheric variables as external forcing. All spatial data is downsampled to a 24-hour temporal resolution (daily mean for ocean and 12:00 UTC for forcings) and 0.5° spatial resolution. Further details on data processing and variables are available in the Appendix.

\paragraph{Baselines.} We compare \method{} against three representative models: 
(1) \textbf{FourCastNet}~\cite{pathak2022FourCastNet}, a widely recognized atmospheric science foundation model based on Fourier Neural Operators.
(2) \textbf{CirT}~\cite{liu2025cirt}, a state-of-the-art atmospheric forecasting model designed for the S2S scale.
(3) \textbf{WenHai}~\cite{cui2025forecasting}, a state-of-the-art ML-based ocean model that has demonstrated superior performance over traditional numerical models and ML-based ocean foundation models.
To ensure a fair comparison, all models, except for the officially released WenHai results, are retrained under our unified framework.

\paragraph{Evaluation Metrics.} We use four metrics to assess performance: (1) \textbf{Root Mean Square Error (RMSE)} to measure the average prediction error, and (2) \textbf{Anomaly Correlation Coefficient (ACC)} to evaluate the model's ability to capture deviations from the climatological mean (i.e., anomalies). Additionally, we use (3) \textbf{Critical Success Index (CSI)}, and (4) \textbf{Symmetric Extremal Dependence Index (SEDI)} to evaluate the model's performance on extreme events.

\subsection{Main Results and Analysis}

\paragraph{\method{} sets a state-of-the-art in long-term simulation.}
We first evaluate \method{} in an idealized simulation setting using ground truth atmospheric forcings. As shown in \textbf{Table~\ref{tab:mainres}}, \method{} consistently outperforms all baselines. At a 60-day lead time, it reduces RMSE by \textbf{$13.3$\%} and improves ACC by \textbf{$12.0$\%} over the strongest baseline, WenHai. This superior performance is visually confirmed in \textbf{Figure~\ref{fig:visual}}, where \method{} is the only model to maintain physical consistency and avoid the error cascades that cause other models to collapse. The significantly lower error growth rate, detailed in \textbf{Figure~\ref{fig:rmse_acc}}, further underscores its long-term stability. The poor performance of powerful atmospheric models like CirT ($60$-day ACC of $0.0081$) also highlights the unique challenges of ocean simulation and validates our specialized design.

\begin{figure*}[t]
  \centering
  \includegraphics[width=\linewidth]{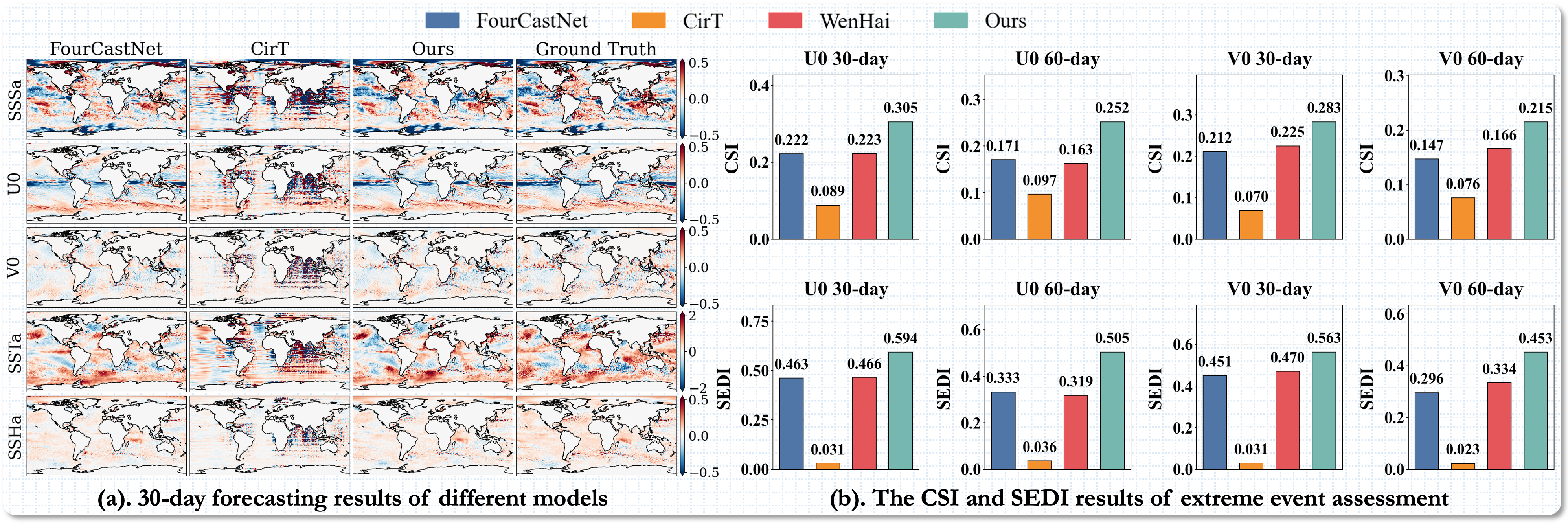}
  \vspace{-20pt}
  \caption{
    Performance on forecasting and extreme event assessment. 
    \textbf{(a)} Visualizations of 30-day forecasts. Our model's outputs align closely with the ground truth, while baseline models exhibit significant artifacts. 
    \textbf{(b)} CSI and SEDI scores for extreme surface current events. \method{} achieves higher scores than baselines at both 30 and 60-day lead times.
  }
  \label{fig:forecasting_and_extremes}
\vspace{-10pt}
\end{figure*}

\paragraph{Robust performance is maintained in ocean forecasting.}
To assess practical applicability, we drive the ocean models with atmospheric \textit{forecasts} from OneForecast, which introduces real-world uncertainty. As common atmospheric forecast models don't provide all forcings required by WenHai at the S2S scale, we can't assess its results. Even in this more challenging scenario, \textbf{Table~\ref{tab:mainres_forecasting}} shows that \method{} retains its lead, achieving significantly better RMSE ($0.68$) and ACC ($0.48$) at a $35$-day lead time. This demonstrates the robustness of \method{} for real-world applications, further supported by the high-fidelity visuals in \textbf{Figure~\ref{fig:forecasting_and_extremes}(a)}.

\paragraph{\method{} excels at capturing high-impact extreme events.} Beyond average metrics, a model's skill in predicting extreme events is critical. We use CSI and SEDI to evaluate predictions of extreme surface currents. As shown in \textbf{Figure~\ref{fig:forecasting_and_extremes}(b)}, \method{} achieves higher scores than all baselines at both 30 and 60-day lead times. This enhanced capability to capture potentially hazardous dynamics demonstrates its value for critical applications like risk assessment.

\begin{table}[htbp]
  \centering
  \vspace{-5pt}
  \setlength{\tabcolsep}{8pt}
  \begin{tabular}{l|cc}
    \toprule
    Variants              & RMSE & ACC \\ 
    \midrule
    \method{} w/o PRC      & 0.6692   & 0.5502  \\
    \method{} w/o PEI      & 0.6865   & 0.5166  \\
    \method{} w/o MANA Sum  & 0.7037   & 0.5359  \\
    \method{} w/o MANA Mean & 0.6978   & 0.5160  \\ 
    \midrule
    \method{}              & \textbf{0.6009}   & \textbf{0.6058}  \\ 
    \bottomrule
   
  \end{tabular}
  \vspace{-5pt}
   \caption{Ablation Studies on model design, the best results are in \textbf{bold}.}
  \label{tab:ablation_mdoel}
   \vspace{-17pt}
\end{table}

\begin{table*}
    
    \small
    \vskip 0.10in
    \centering
    \vspace{-5pt}
        \begin{sc}
            \renewcommand{\multirowsetup}{\centering}
            \setlength{\tabcolsep}{11pt} 
            \begin{tabular}{l|cc|cc|cc}
            \toprule
            \multirow{2}{*}{VARIANTS}         & \multicolumn{2}{c|}{23 Layers S} & \multicolumn{2}{c|}{23 Layers T} & \multicolumn{2}{c}{SSH} \\ \cmidrule{2-7} 
                                              & RMSE            & ACC            & RMSE            & ACC            & RMSE        & ACC       \\ \midrule
            \method{} w/o Subtract Climatology & 0.0515              & 0.1066             & 0.1518              & 0.4532             & 0.2427          & 0.5737        \\
            \method{} w Subtract Climatology   & \textbf{0.0105}             & \textbf{0.6412}             & \textbf{0.0985}              & \textbf{0.6507}             & \textbf{0.1375}          & \textbf{0.6690}        \\ \bottomrule
            \end{tabular}
        \end{sc}
        \vspace{-5pt}
        \caption{Ablation Studies on whether subtracting the climatology from periodicity variables will improve the simulation performance, the best results are in \textbf{bold}.}
        \vspace{-15pt}
        \label{tab:ablation_var}
        
\end{table*}

\begin{table}[t]
\vspace{-10pt}
    \vskip 0.13in
    \centering
    \begin{small}
        \begin{sc}
            \renewcommand{\multirowsetup}{\centering}
            \setlength{\tabcolsep}{2.4pt} 
           \begin{tabular}{l|cccccc}
                \toprule
                \multirow{3}{*}{Models} & \multicolumn{6}{c}{Metrics}                                                                                                             \\ \cmidrule{2-7} 
                                        & \multicolumn{2}{c|}{15-day}                        & \multicolumn{2}{c|}{25-day}                        & \multicolumn{2}{c}{35-day}    \\ \cmidrule{2-7} 
                                        & RMSE          & \multicolumn{1}{c|}{ACC}           & RMSE          & \multicolumn{1}{c|}{ACC}           & RMSE          & ACC           \\ \midrule
                FourCastNet             & 0.55          & \multicolumn{1}{c|}{0.68}          & 0.68          & \multicolumn{1}{c|}{0.47}          & 1.30          & 0.19          \\
                CirT                    & 1.36          & \multicolumn{1}{c|}{0.14}          & 2.08          & \multicolumn{1}{c|}{0.05}          & 2.61          & 0.02          \\ \midrule
                Ours                    & \textbf{0.52} & \multicolumn{1}{c|}{\textbf{0.73}} & \textbf{0.62} & \multicolumn{1}{c|}{\textbf{0.59}} & \textbf{0.68} & \textbf{0.48} \\ \bottomrule
                \end{tabular}
        \end{sc}
    \end{small}
    \vspace{-5pt}
    \caption{In the real-world global ocean forecasting task, we compare the performance of our \method{} with 2 baselines. The average results for all 93 ocean variables of RMSE and ACC are recorded. The best results are in \textbf{bold}.}
    \label{tab:mainres_forecasting}
    \vspace{-20pt}
\end{table}

\subsection{Ablation Studies}
To understand the driving factors behind \method{}'s superior performance and to validate the effectiveness of each of our design choices, we conduct detailed ablation studies. We primarily investigate two aspects: the importance of the input representation (i.e., subtracting climatology) and the contribution of each component in our core architecture.

\paragraph{Subtracting climatology priors is critical for performance.}
As mentioned in our methodology, we argue that focusing the model on learning ``anomalies" rather than the full state is crucial for capturing slow-changing dynamics. To verify this, we train a variant of \method{} without subtracting the climatological mean. The results, shown in \textbf{Table~\ref{tab:ablation_var}}, are striking. For sea salinity (S) across 23 depth layers, the model with climatology priors (W Subtract Climatology) achieves an ACC of 0.6412, whereas the model without them (W/O Subtract Climatology) only reaches 0.1066—a performance gap of over \textbf{5-fold}. This result provides strong evidence that incorporating physical priors like climatology into the input representation is a \textbf{prerequisite} for enabling the model to effectively learn fine-grained physical signals and achieve high-fidelity simulations.

\paragraph{Each core architectural component is indispensable.}
We evaluate the individual contributions of \method{}'s core components by removing them one by one. The results, summarized in \textbf{Table~\ref{tab:ablation_mdoel}}, show that removing any single component leads to a significant degradation in performance:

\noindent \ding{224} \textbf{\textit{Without Progressive Residual Correction} (w/o PRC):} Replacing the multi-stage PRC framework with a single model increases RMSE from 0.6009 to 0.6692, confirming that decomposing the complex prediction task into a series of simpler residual correction steps is vital for suppressing error accumulation and ensuring long-term stability.

\noindent \ding{224} \textbf{\textit{Without Physics-inspired Edge-Interactions} (w/o PEI):} Removing PEI, thereby reverting to a traditional, physics-agnostic message passing scheme, caused the ACC to drop from 0.6058 to 0.5166. This confirms that explicitly modeling physical processes like gradients and couplings in message passing improves the model's physical consistency.

\noindent \ding{224} \textbf{\textit{Multi-scale Adaptive Node Aggregation} (w/o MANA Sum/Mean):} Removing either the sum or the mean aggregation path degrades performance. This confirms the necessity of our adaptive aggregation mechanism, which allows the model to flexibly handle ocean dynamics at different scales.

In summary, these ablation studies confirms that the success of \method{} is not coincidental but is the result of a synergistic combination of its carefully designed components from the input representation and macroscopic framework to the microscopic interaction mechanisms.

\section{Conclusion}

In this paper, we introduce \method{}, a neural operator framework designed to solve the problem of long-term stability by simulating slow-changing physical systems. Its core innovations, a Progressive Residual Correction Framework and a Physics-Guided Graph Network, synergistically improve long-term stability and physical consistency. In the challenging task of global S2S ocean simulation, experiments demonstrate that \method{} significantly outperforms state-of-the-art models in forecast accuracy, long-term stability, and the ability to capture extreme events. Future work could explore integrating strict physical conservation laws into the loss function and reducing computational costs.

\clearpage

\section*{Acknowledgements}
This work was supported by the National Natural Science Foundation of China (42125503, 42430602).
\bibliography{aaai2026}

\clearpage

\clearpage 

\section{A ~~~Data Details}
\label{appendix:data}
\subsection{A.1 ~~~Dataset}
We conduct the experiments on GlORYS12 reanalysis data, offering daily mean data covering latitudes between -80° and 90° from 1993 to the present, which can be downloaded from \url{https://data.marine.copernicus.eu}. The subset we use includes years from 1993 to 2020, which is 1993-2017 for training, 2018-2019 for validating, and 2020 for testing. We use 4 depth level ocean variables (each with 23 depth levels, corresponding to 0.49 m, 2.65 m, 5.08 m, 7.93 m, 11.41 m, 15.81 m, 21.60 m, 29.44 m, 40.34 m, 55.76 m, 77.85 m, 92.32 m, 109.73 m, 130.67 m, 155.85 m, 186.13 m, 222.48 m, 266.04 m, 318.13 m, 380.21 m, 453.94 m, 541.09 m and 643.57 m), Sea salinity (S), Sea stream zonal velocity (U), Sea stream meridional velocity (V), Sea temperature (T), and 1 surface level variable Sea surface height (SSH). To improve computational efficiency, we use bilinear interpolation to downsampling them to 1/2 degree (H=361, W=720). And to better adapt to the input of different architecture models, we use the data with size 360 × 720. 4 atmosphere variables from ERA5 are used as forcing, which include 10 metre u wind component (U10M), 10 metre v wind component (V10M), 2 metre temperature (T2M), and mean sea level pressure (MSLP). The ERA5 data can be downloaded from \url{https://cds.climate.copernicus.eu}, the official website of Climate Data Store (CDS). All the data we used are shown in Table \ref{tab:appendix_data}.

\subsection{A.2 ~~~Data Preprocessing}
Different ocean and forcing variables exhibit substantial variations in magnitude. To enable the model to concentrate on accurate simulation rather than learning the inherent magnitude discrepancies among variables, we normalize the input data prior to model ingestion. Specifically, for ocean variables, we compute the mean and standard deviation from the training dataset spanning the years 1993 to 2017. For forcing variables, we calculate these statistics from an extended dataset covering the period 1959 to 2017. Each variable thus possesses a dedicated mean and standard deviation. Before inputting data into the model, we normalize the data by subtracting the corresponding mean and dividing the respective standard deviation. For the `nan' values of land, we fill them with zero before inputting the data into the model. 

\begin{table*}[ht]
\centering
\setlength{\tabcolsep}{7pt}
\begin{tabular}{ccccccc}
\toprule
\textbf{Type}        & \textbf{Full name}                 & \textbf{Abbreviation} & \textbf{Layers} & \textbf{Time}      & \textbf{Dt} & \textbf{Spatial Resolution}\\ \midrule
Atmosphere & 10 metre u wind component     & U10M         & 1      & 1993-2020 & 24h             & 0.5°               \\
Atmosphere & 10 metre v wind component     & V10M         & 1      & 1993-2020 & 24h             & 0.5°               \\
Atmosphere & 2 metre temperature           & T2M          & 1      & 1993-2020 & 24h             & 0.5°               \\
Atmosphere & Mean sea level pressure       & MSLP         & 1      & 1993-2020 & 24h             & 0.5°               \\ \midrule
Ocean     & Sea salinity                  & S            & 23     & 1993-2020 & 24h             & 0.5°               \\
Ocean     & Sea stream zonal velocity     & U         & 23     & 1993-2020 & 24h             & 0.5°               \\
Ocean     & Sea stream meridional velocity & V         & 23     & 1993-2020 & 24h             & 0.5°               \\
Ocean     & Sea temperature               & T         & 23     & 1993-2020 & 24h             & 0.5°               \\
Ocean     & Sea surface height            & SSH          & 1      & 1993-2020 & 24h             & 0.5°               \\ \midrule
Static      & Land-sea mask                 & LSM          & ---    & ---       & ---             & 0.5°               \\ \bottomrule
\end{tabular}
\caption{The data details in this work.}
\label{tab:appendix_data}
\vspace{-9pt}
\end{table*}

\section{B ~~~Details of Physics-Guided Graph Network}

\subsection{B.1 ~~~Multi-scale Graph Encoder}
We employ the multi-scale graph encoder to map the data from lat-lon grids into a graph structure. The multi-scale graph structure can be defined as:
\begin{equation}
\mathcal{G}=\left(\mathcal{V}^G, \mathcal{V}, \mathcal{E}_m, \mathcal{E}^{\mathrm{G} 2 \mathrm{M}}, \mathcal{E}^{\mathrm{M} 2 \mathrm{G}}\right),
\end{equation}
where, ${\mathcal{V}}^{G}$ represents the set of lat-lon grid nodes, with a total of $N = H \times W$ nodes; $\mathcal{V}$ represents the mesh nodes; $\mathcal{E}_m$ denotes the multi-scale edges, which represents the edge with different lengths; $\mathcal{E}^{\mathrm{G} 2 \mathrm{M}}$ and $\mathcal{E}^{\mathrm{M} 2 \mathrm{G}}$ are the unidirectional edges that connect lat-lon grid nodes and mesh nodes. All scales share the same set of nodes $\mathcal{V}$ but with multi-level edges. Then, we apply an MLP to map the data to the latent space, which can be defined as:
\begin{equation}
    (\mathcal{V}_f^G, h, \mathcal{E}_f, \mathcal{E}^{\mathrm{G} 2 \mathrm{M}}_f, \mathcal{E}^{\mathrm{M} 2 \mathrm{G}}_f) = \mathrm{MLP}(X_t, \mathcal{V}, \mathcal{E}_m, \mathcal{E}^{\mathrm{G} 2 \mathrm{M}}, \mathcal{E}^{\mathrm{M} 2 \mathrm{G}}),
\end{equation}
\begin{equation}
    MLP = \mathcal{N}(\sigma(\mathcal{L}(\cdot))),
\end{equation}
where, $\mathcal{L(\cdot)}$ is the linear function. $\sigma(\cdot)$ is the SiLU activation function, $\mathcal{N}(\cdot)$ is the LayerNorm function. $X_t$, $\mathcal{V}$, $\mathcal{E}$, $\mathcal{E}^{\mathrm{G} 2 \mathrm{M}}$, and $\mathcal{E}^{\mathrm{M} 2 \mathrm{G}}$ are embedded features of grid nodes, mesh nodes, mesh edges, grid to mesh edges, and mesh to grid edges. Then, we update the grid2mesh edge features using information from the adjacent nodes:
\begin{equation}
     {\mathcal{E}^{\mathrm{G} 2 \mathrm{M}}_f}^{\prime} = \mathrm{ESMLP}(\mathcal{V}_f^G, h, \mathcal{E}^{\mathrm{G} 2 \mathrm{M}}_f),
\end{equation}
where, $\mathrm{ESMLP}(\cdot)$ is the Edge Sum MLP~\cite{pfaff2020learning}. The mesh node features are updated by an MLP:
\begin{equation}
    h^{\prime} = \mathrm{MLP_{e1}}(h, \sum{\mathcal{E}^{\mathrm{G}2\mathrm{M}}_f}^{\prime}),
\end{equation}
where, $\sum \mathcal{E}_f^{\mathrm{G} 2 \mathrm{M}^{\prime}}$ are the edges that arrives at mesh node. The grid node features are updated using another MLP:
\begin{equation}
\mathcal{V}_f^{G^{\prime}}=\operatorname{MLP}_{e2}\left(\mathcal{V}_f^G\right).
\end{equation}
Finally, the residual connections are applied to update $\mathcal{E}^{\mathrm{G} 2 \mathrm{M}}_f$, $h$, and $\mathcal{V}_f^G$ again:
\begin{equation}
    {\mathcal{E}^{\mathrm{G} 2 \mathrm{M}}_f} = {\mathcal{E}^{\mathrm{G} 2 \mathrm{M}}_f} + {\mathcal{E}^{\mathrm{G} 2 \mathrm{M}}_f}^{\prime},
    h = h + h^{\prime},
    \mathcal{V}_f^{G} = \mathcal{V}_f^{G} + {\mathcal{V}_f^{G}}^{\prime}    .
\end{equation}

\subsection{B.2 ~~~Physics-guided Adaptive Graph Messaging}
To make the message passing more consistent with the evolution process of the ocean dynamics system, we propose a module called Physics-guided Adaptive Messaging (PAGM). PAGM includes a
physics-inspired edge-interactions module and a multi-scale adaptive node aggregation module. In our setup, 16 PAGM are applied to conduct massaging.

\subsubsection{Physics-Inspired Edge-Interactions.} In ocean simulation, ocean currents, temperature, salinity, and other factors interact with each other, and these interactions span different scales and layers. Unlike traditional MLP-based update strategy, which update edges by simply concatenating edges and nodes, we introduce a physics-inspired edge-interactions module to more accurately simulate the dynamic relationships in the ocean system. For the sender node feature $\mathbf{h}_{s(i)}$ and receiver nodes feature $\mathbf{h}_{r(i)}$, we first calculate their mutual interaction:
\begin{equation}
    \mathbf{h}_{d(i)} = \mathbf{h}_{s(i)} - \mathbf{h}_{r(i)},
\end{equation}
where, $\mathbf{h}_{d(i)}$ represents the variation in features between neighboring nodes, capturing the gradient changes of quantities such as flow velocity and salinity in the ocean. For example, when simulating changes in ocean currents, the velocity differences can indicate the energy transfer and dynamic characteristics of different ocean regions.
\begin{equation}
    \mathbf{h}_{mp(i)} = \mathbf{h}_{s(i)}*\mathbf{h}_{r(i)},
\end{equation}
where, $\mathbf{h}_{mp(i)}$ represents the multiplicative coupling relationship between nodes, similar to the coupling phenomenon between temperature and salinity in the ocean. The temperature and salinity of seawater together determine its density, which is crucial for modeling deep ocean currents.
\begin{equation}
    \mathbf{h}_{cos(i)} = cos(\mathbf{h}_{s(i)}, \mathbf{h}_{r(i)}),
\end{equation}
where, $\mathbf{h}_{cos(i)}$ is the cosine similarity between different nodes.  For the ocean systems, it can be used to identify similar water masses or marine subsystems. For instance, when the changes in seawater temperature and salinity become consistent, it may indicate that these two ocean regions are influenced by similar climate systems, allowing information to be transmitted through similarity and improving the accuracy of the simulation. We define a function $\mathcal{R}(\cdot)$ to fuse mesh node features and the features produced by the interaction:
\begin{equation}
    \mathcal{R}(x) = \mathcal{\sigma}(\mathcal{N}(\mathcal{L}(\mathcal{C}(x)))),
\end{equation}
where, $\mathcal{C}$ represents the concatenate function. Then, the fused mesh node features can be obtained by:
\begin{equation}
    \mathbf{h}_{s(i)} = \mathcal{R}(\mathbf{h}_{s(i)}, \mathbf{h}_{mp(i)}, \mathbf{h}_{mp(i)}, \mathbf{h}_{cos(i)}),
\end{equation}
\begin{equation}
    \mathbf{h}_{r(i)} = \mathcal{R}(\mathbf{h}_{r(i)}, \mathbf{h}_{mp(i)}, \mathbf{h}_{mp(i)}, \mathbf{h}_{cos(i)}).
\end{equation}
Subsequently, the edge features are updated using the information from adjacent nodes:
\begin{equation}
    \mathcal{E}_f^{\prime}=\mathbf{W}_e \mathcal{E}_f^{},
    h_s^{\prime}=\mathbf{W}_s h_s,
     h_r^{\prime}=\mathbf{W}_r h_r+\mathbf{b}_r,
\end{equation}
\begin{equation}
    \mathbf{h}_{\mathrm{sum}}=\mathcal{E}_f^{\prime}+h_s^{\prime}+h_r^{\prime},
\end{equation}
\begin{equation}
    {\mathcal{E}_f}^{\prime}=\mathcal{N}\left(\mathbf{W}{\cdot} \sigma\left(h_{sum}\right)+\mathbf{b}\right),
\end{equation}
where, $\mathbf{W}_e$, $\mathbf{W}_s$, $\mathbf{W}_r$ are the linear transformation matrix of edge features, send node feature, and receive node features. $\mathbf{W}$ is the linear transformation matrix of output layer. $b_r$ and $b$ are the bias.
Finally, the edge features are updated using the residual connection:
\begin{equation}
    {\mathcal{E}_f} = {\mathcal{E}_f}^{\prime} + {\mathcal{E}_f}.
\end{equation}

\subsubsection{Multi-scale Adaptive Node Aggregation.}
Unlike traditional node update strategies, which typically sum or average the features of neighboring nodes, we consider the movements of the ocean at different scales. To achieve adaptive adjustment of aggregation methods across different scales, we introduce a multi-scale adaptive node aggregation module that dynamically adjusts the weights of the aggregation strategy. For rapid local ocean dynamics with smaller scales, the gating mechanism automatically tends to favor sum aggregation, capturing the local rapid dynamic changes quickly and accurately. Conversely, when the model identifies that the motion scale around the node is larger and the changes are slower, it tends to prefer mean aggregation, which achieves a smoother and more robust representation of the large-scale ocean state.
Specifically, in the ocean system, meso- and small-scale movements (such as eddies, localized strong upwelling, coastal currents, etc.) exhibit distinct characteristics of localized energy concentration and rapid dynamic changes. In these scenarios, the influence of adjacent regions on a node is often cumulative, such as the concentrated propagation of local heat, momentum, or salt flux. In such cases, the sum aggregation strategy, by directly adding the contributions from surrounding nodes, more accurately reflects the rapid evolution and locally concentrated dynamic features, which can be expressed as:
\begin{equation}
h_{sum}^{\prime}=\mathrm{MLP}_{sum}(h, \sum \mathcal{E}_f^{\prime}).
\end{equation}
For large-scale, slowly varying ocean movements (such as the global ocean circulation, thermohaline circulation, and long-term ocean current trends), the differences between nodes and the extent of local fluctuations are smaller, and information transfer between neighboring nodes tends to be in a relatively steady and averaged state. In this case, the mean aggregation strategy can more stably capture long-term trends and the overall balanced state, avoiding the over-influence of short-term local disturbances on the large-scale state, which can be expressed as:
\begin{equation}
h_{mean}^{\prime}=\mathrm{MLP}_{mean}(h, \sum \mathcal{E}_f^{\prime}).
\end{equation}
Then, we use $h_{sum}^{\prime}$ and $h_{mean}^{\prime}$ to calculate gating coefficient:
\begin{equation}
    \gamma = \mathrm{MLP_g}(\mathcal{C}(h_{sum}^{\prime}, h_{mean}^{\prime})).
\end{equation}
Finally, an MLP and a residual connection are used to update the node features:
\begin{equation}
    h^{\prime} = \mathrm{MLP_{node}}(\gamma{\cdot}h_{sum}^{\prime} + (1-\gamma){\cdot}h_{mean}^{\prime}).
\end{equation}

\subsubsection{B.3 ~~~Decoder}
The role of the decoder is to decode the information from the latent space to the lat-lon grid. We first use the information from the adjacent nodes to update mesh2grid features:
    \begin{equation}
        \mathcal{E}_f^{\mathrm{M} 2 \mathrm{G}^{\prime}}=\operatorname{ESMLP}\left(\mathcal{V}_f^G, h, \mathcal{E}_f^{\mathrm{M} 2 \mathrm{G}}\right).
    \end{equation}
    Then, we update the grid node features using the information of edges that arrive at the grid nodes:
    \begin{equation}
        {\mathcal{V}_f^G}^{\prime} = {\mathrm{MLP}}_{d1}(\mathcal{V}_f^G, \sum{\mathcal{E}^{\mathrm{M} 2 \mathrm{G}}_f}^{\prime}).
    \end{equation}
    Subsequently, we apply residual connection to update grid node features:
    \begin{equation}
    \mathcal{V}_f^{G}=\mathcal{V}_f^G+\mathcal{V}_f^{G^{\prime}}.
    \end{equation}
    Finally, we use an MLP to predict the next step results:
    \begin{equation}
    \hat{O}_{t+1}=\operatorname{MLP}_{d2}\left(\mathcal{V}_f^{G}\right).
    \end{equation}

\section{C ~~~Experiments Details}
\subsection{C.1 ~~~Evaluation Metrics}
We utilize four metrics, RMSE (Root Mean Square Error), ACC (Anomalous Correlation Coefficient), CSI (Critical Success Index), and SEDI (Symmetric Extremal Dependence Index) to evaluate the simulation performance, which can be defined as:
\begin{equation}
    \small
    \operatorname{RMSE}(\mathcal{J}, t) = \sqrt{\frac{\sum\limits_{i=1}^{N_{\text{lat}}} \sum\limits_{j=1}^{N_{\text{lon}}} L(i) \left( \hat{\mathbf{A}}_{ij,t}^\mathcal{J} - \mathbf{A}_{ij,t}^\mathcal{J} \right)^2}{N_{\text{lat}} \times N_{\text{lon}}}},
\end{equation}
\begin{equation}
    \small
    \operatorname{ACC}(\mathcal{J}, t) = \frac{\sum\limits_{i=1}^{N_{\text{lat}}} \sum\limits_{j=1}^{N_{\text{lon}}} L(i) \hat{\mathbf{A'}}_{ij,t}^\mathcal{J} \mathbf{A'}_{ij,t}^\mathcal{J}}{\sqrt{\sum\limits_{i=1}^{N_{\text{lat}}} \sum\limits_{j=1}^{N_{\text{lon}}} L(i) \left( \hat{\mathbf{A'}}_{ij,t}^\mathcal{J} \right)^2 \times \sum\limits_{i=1}^{N_{\text{lat}}} \sum\limits_{j=1}^{N_{\text{lon}}} L(i) \left( \mathbf{A'}_{ij,t}^\mathcal{J} \right)^2}},
\end{equation}
where $\mathbf{A}_{i, j, t}^\mathcal{K}$ represents the value of variable $\mathcal{J}$ at horizontal coordinate $(i, j)$ and time t. Latitude-dependent weights are defined as $L(i)=N_{\text {lat }} \times \frac{\cos \phi_i}{\sum_{i’=1}^{N_{\text {lat}}} \cos \phi_{i’}}$, where $\phi_i$ is the latitude at index i. The anomaly of $A$, denoted as $A'$, is computed as the deviation from its climatology, which corresponds to the long-term mean of the meteorological state estimated from multiple years of training data. RMSE and ACC are averaged across all time steps and over non-NaN spatial grid points, providing summary statistics for each variable $\mathcal{J}$ at a given lead time $\Delta t$.

\begin{equation}
\small
\operatorname{CSI}(\mathcal{J}, t)=\frac{\mathrm{TP}}{\mathrm{TP}+\mathrm{FP}+\mathrm{FN}},
\end{equation}
\begin{equation}
\small
\operatorname{SEDI}(\mathcal{J}, t)=\frac{\log (F)-\log (H)-\log (1-F)+\log (1-H)}{\log (F)+\log (H)+\log (1-F)+\log (1-H)},
\end{equation}
where, true positives (TP) represent the number of instances in which the ocean state is correctly simulated, while false positives (FP) and false negatives (FN) are defined analogously. $F=\frac{\mathrm{FP}}{\mathrm{FP}+\mathrm{TP}}$ is the false alarm rate, and $H=\frac{\mathrm{TP}}{\mathrm{TP}+\mathrm{FN}}$ represents the hit rate. In the simulation task, we report the average results for 240 initial conditions (ICs), and in forecasting, for 50 ICs.

\subsection{C.2 ~~~Model Training}

For the progressive residual correction framework, we apply Q sub-models to progressively capture subtle changes in ocean variables. We train the first sub-model with M steps finetune. For other sub-models, we train them with N steps finetune. In our experiment, we set Q=2, M=6, and N=10 as an example. Specifically, we train the first sub-model, denoted as $\mathcal{F}^{1}_1 $ with 200 epochs using 1-step supervision. The learning rate is set to 1e-3. And we use the cosine annealing scheduler to adjust the learning rate until the model converges. After that, we finetune $\mathcal{F}^{1}_1$ with 10 epochs using 2-step supervision to get $\mathcal{F}^{2}_1$. Then, we finetune $\mathcal{F}^{2}_1$ with 10 epochs using 3-step supervision to get $\mathcal{F}^{3}_1$. The learning rate is set to 1e-6. And we use the cosine annealing scheduler to adjust the learning rate until the model converges. The model $\mathcal{F}^{4}_1$, $\mathcal{F}^{5}_1$, and $\mathcal{F}^{M}_1$ are finetuned same as $\mathcal{F}^{3}_1$. Then we freeze the parameters of $\mathcal{F}^{M}_1$, and the second sub-model $\mathcal{F}^{N}_2$ can be trained in two ways. One way is that we first pre-train $\mathcal{F}^1_2$, and then the $\mathcal{F}^N_2$ can be get step by step. The other way is that we directly train $\mathcal{F}^N_2$. In practice, the model converges after a few epochs, and we stop the training early. Furthermore, the Q-th model $\mathcal{F}^N_Q$ can be trained step by step if a higher accuarcy is required. We train all baseline models and our NeuralOM using the same training framework. We will open-source all of our materials—including all training codes, detailed training procedures, testing codes, pre-trained weights, etc. 

\section{D ~~~Additional Results}
\begin{figure*}[ht]
\centering
\includegraphics[width=0.8\linewidth]{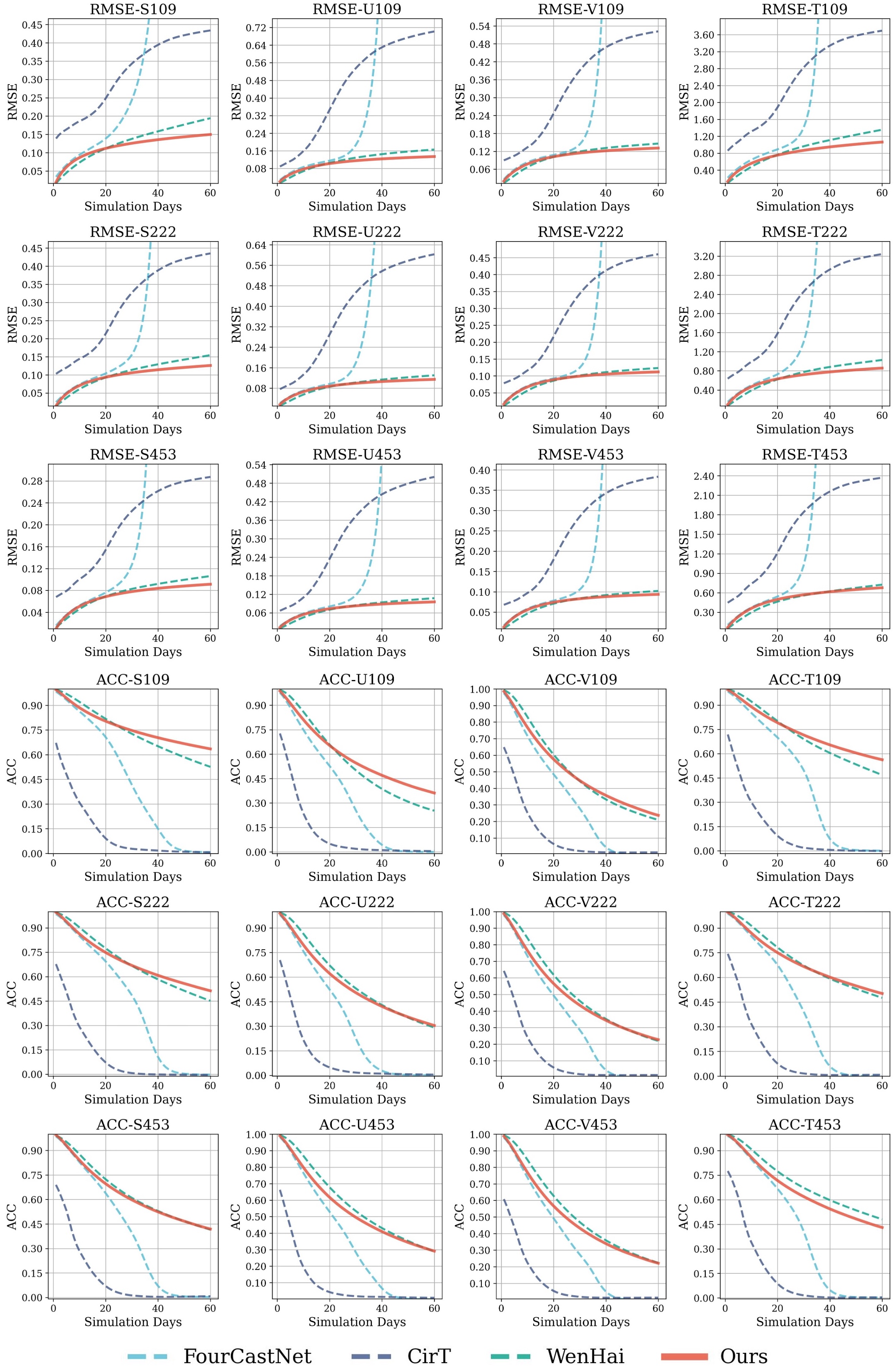}
\caption{The latitude-weighted RMSE (lower is better) and ACC (higher is better) results of several
important ocean variables (depth levels at 109.73 m, 222.48 m, and 453.94 m).}
\label{fig_acc_rmse_appendix}
\end{figure*}

\begin{figure*}[ht]
\centering
\includegraphics[width=0.9\linewidth]{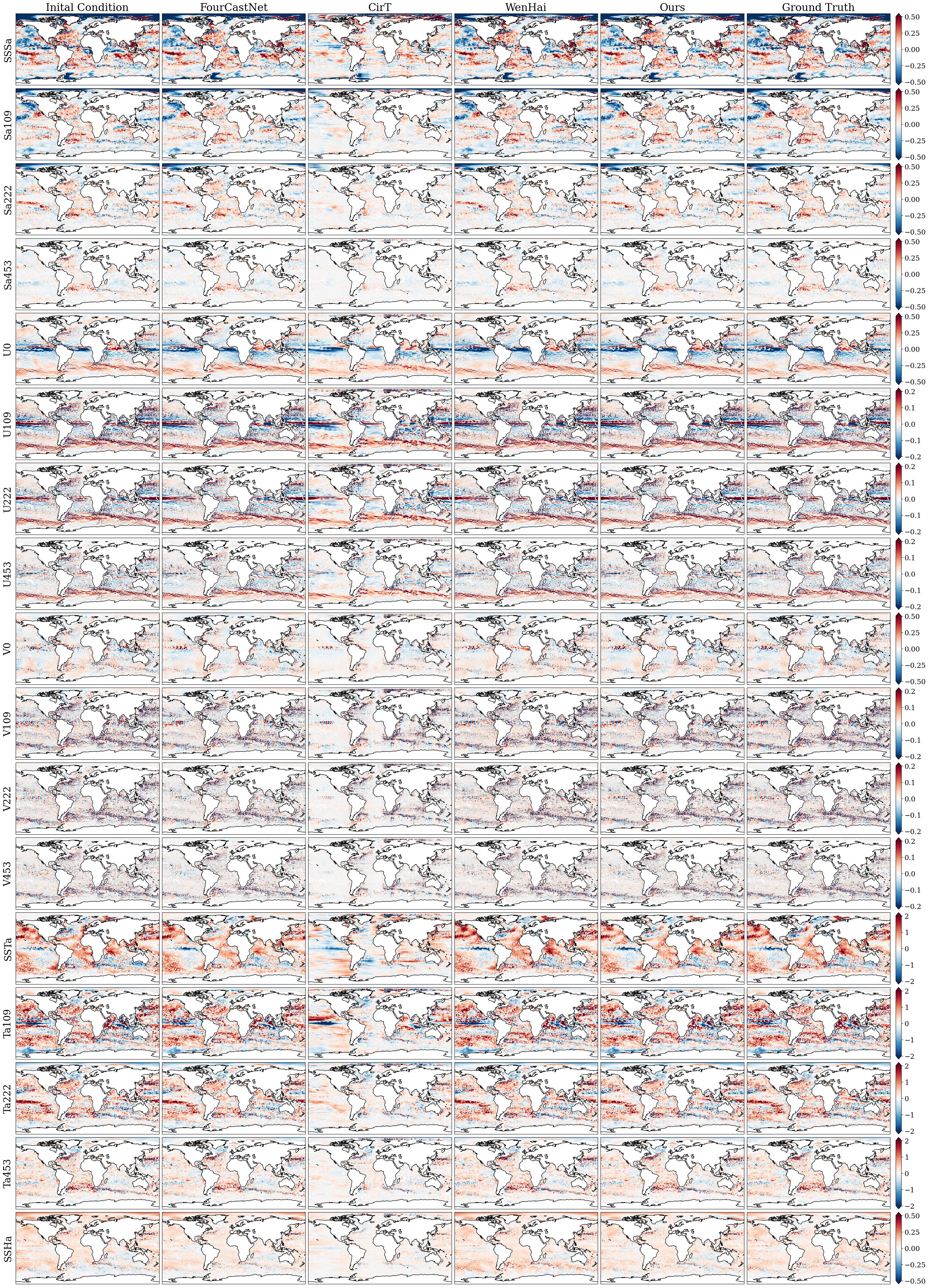}
\caption{10-day simulation results of different models.}
\label{fig_10.0-day}
\end{figure*}

\begin{figure*}[t]
\centering
\includegraphics[width=0.9\linewidth]{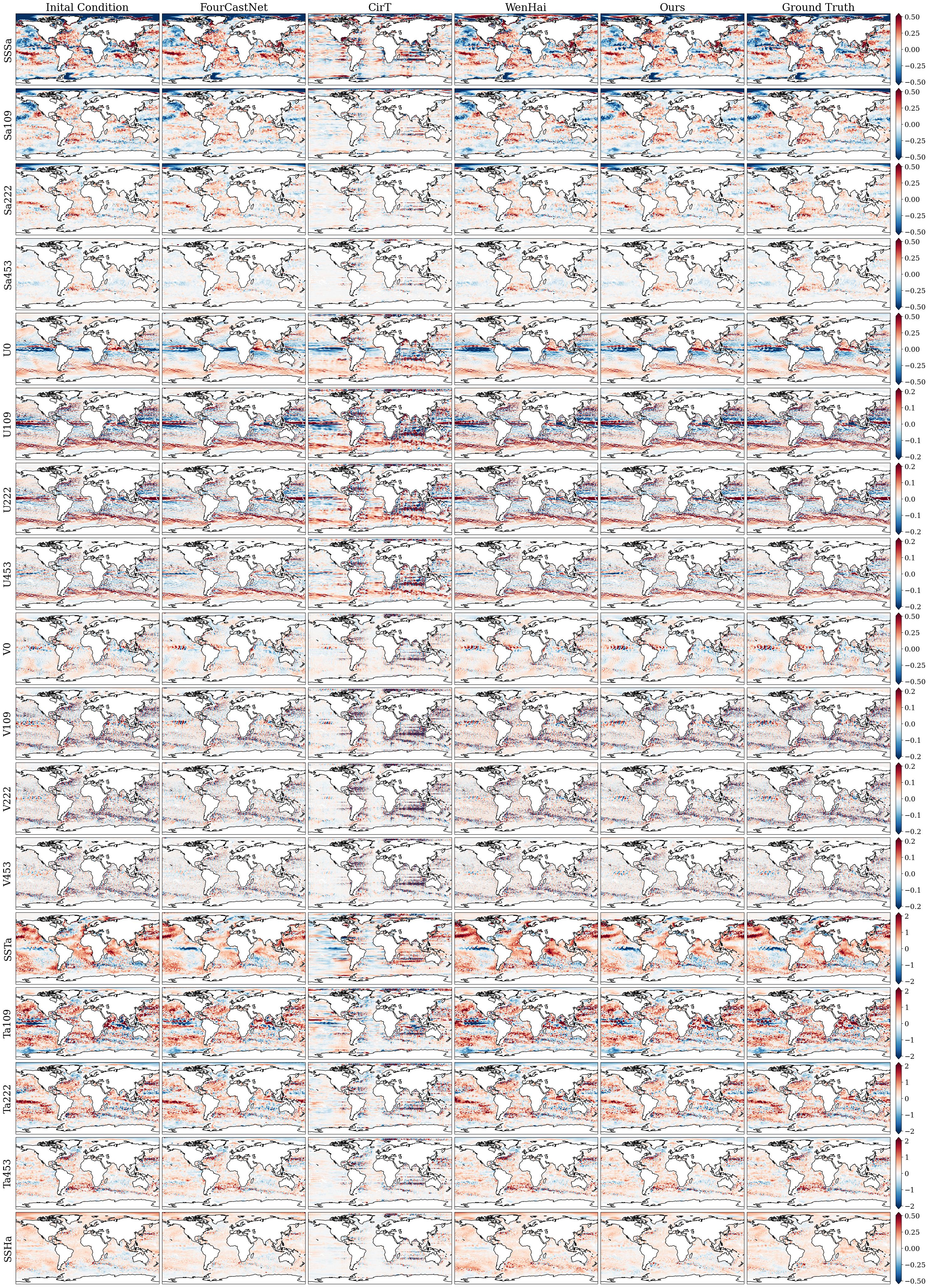}
\caption{20-day simulation results of different models.}
\label{fig_20.0-day}
\end{figure*}

\begin{figure*}[t]
\centering
\includegraphics[width=0.9\linewidth]{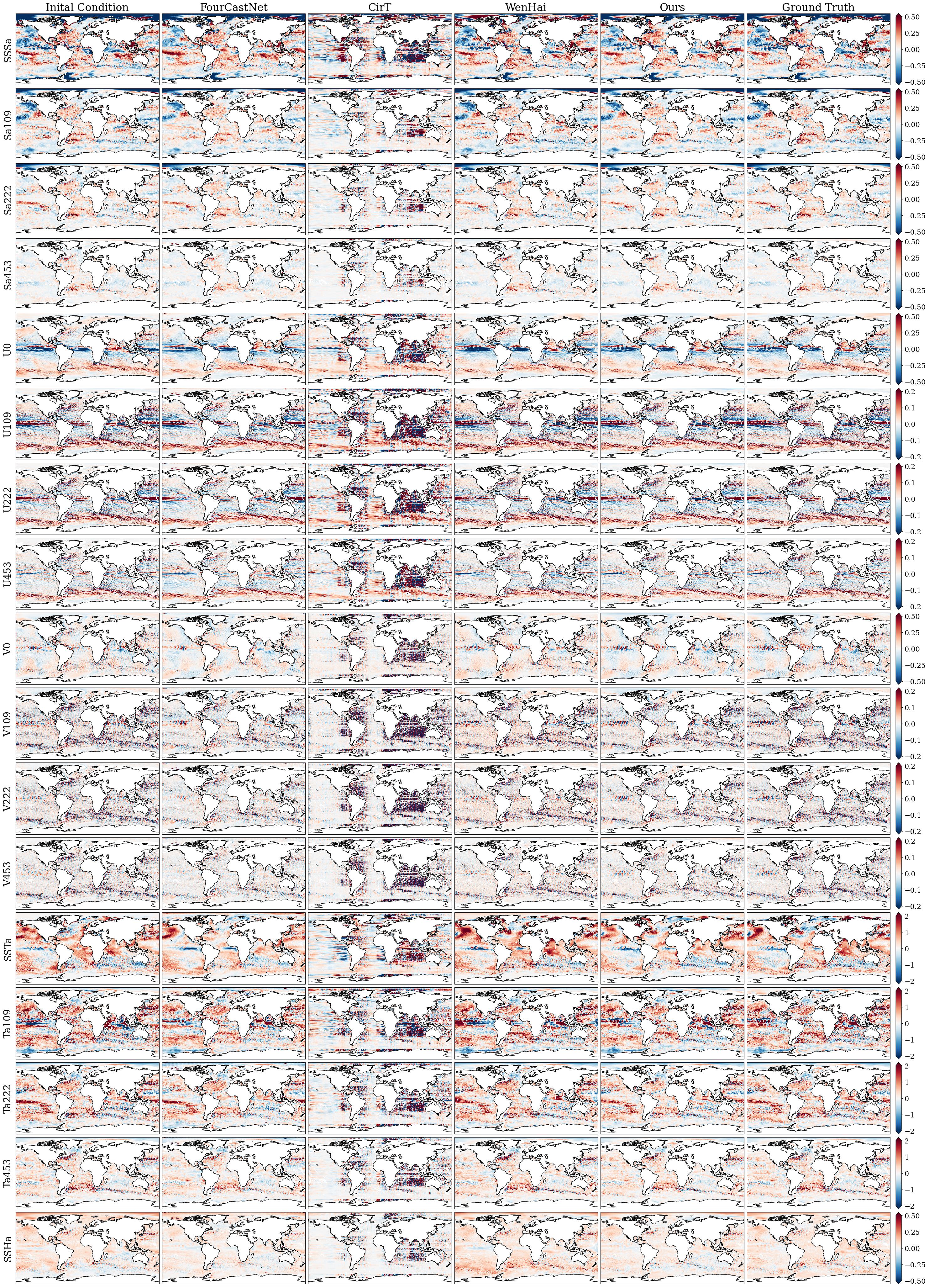}
\caption{30-day simulation results of different models.}
\label{fig_30.0-day}
\end{figure*}

\begin{figure*}[t]
\centering
\includegraphics[width=0.9\linewidth]{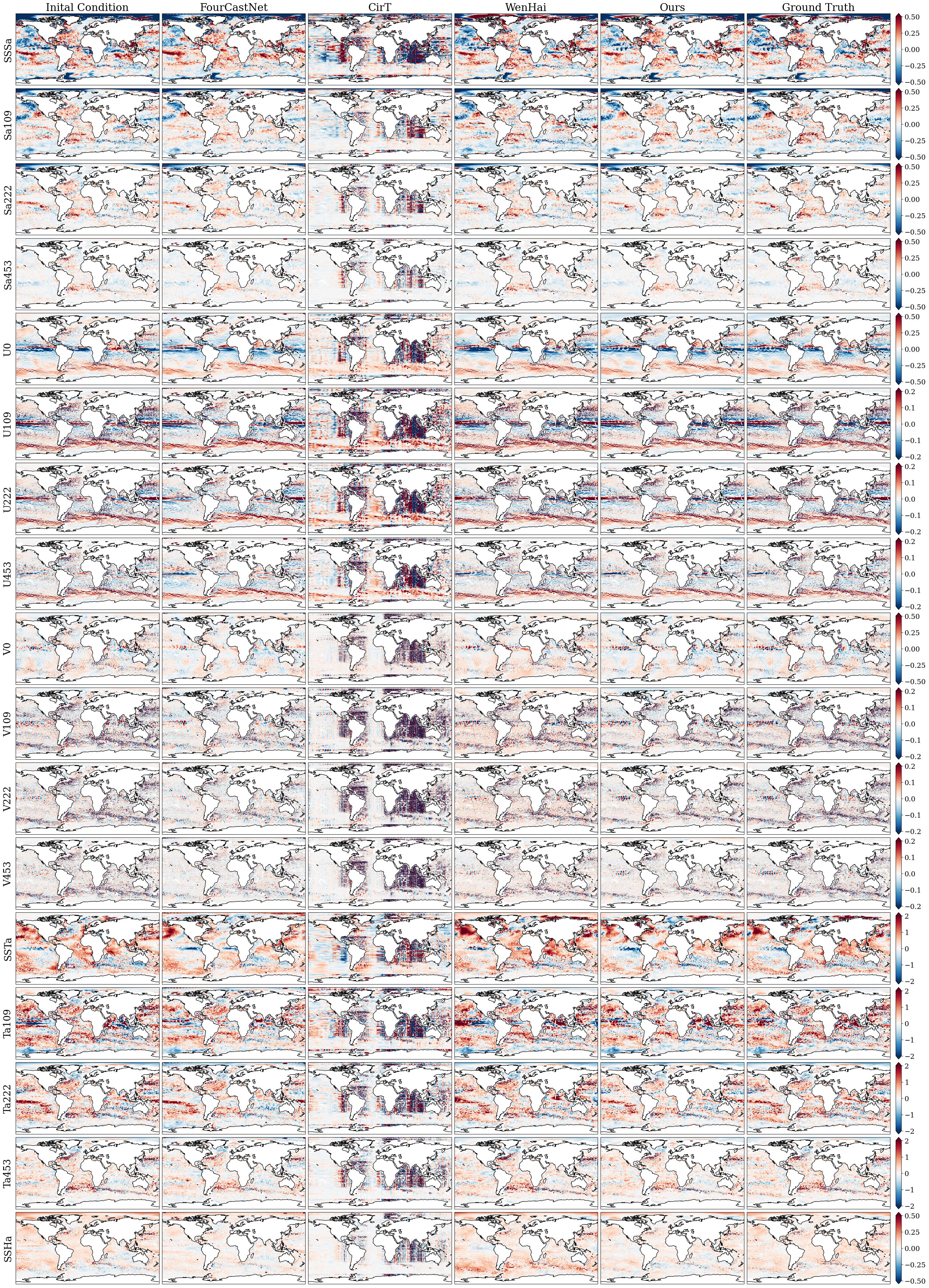}
\caption{40-day simulation results of different models.}
\label{fig_40.0-day}
\end{figure*}

\begin{figure*}[t]
\centering
\includegraphics[width=0.9\linewidth]{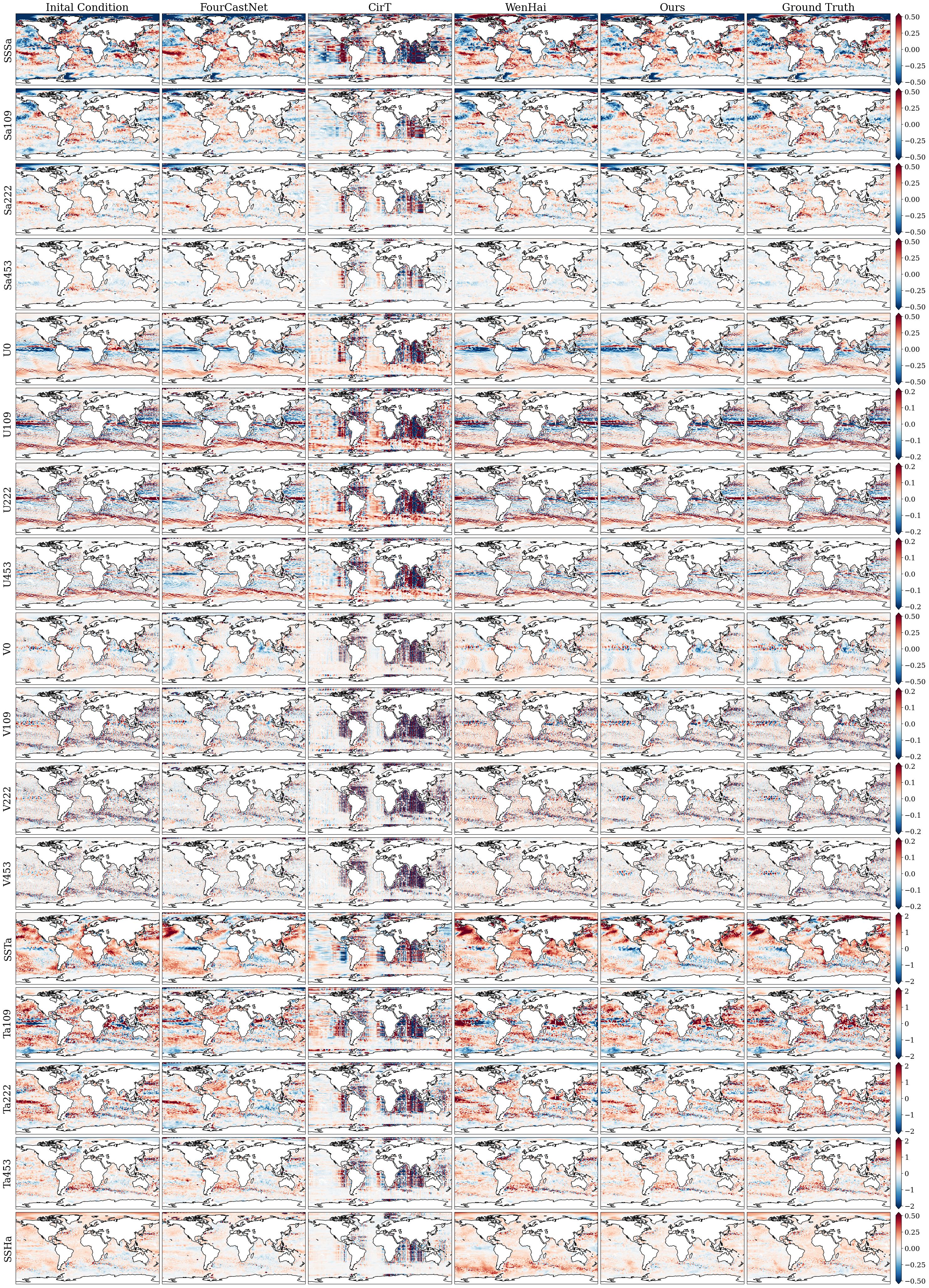}
\caption{50-day simulation results of different models.}
\label{fig_50.0-day}
\end{figure*}

\begin{figure*}[t]
\centering
\includegraphics[width=0.9\linewidth]{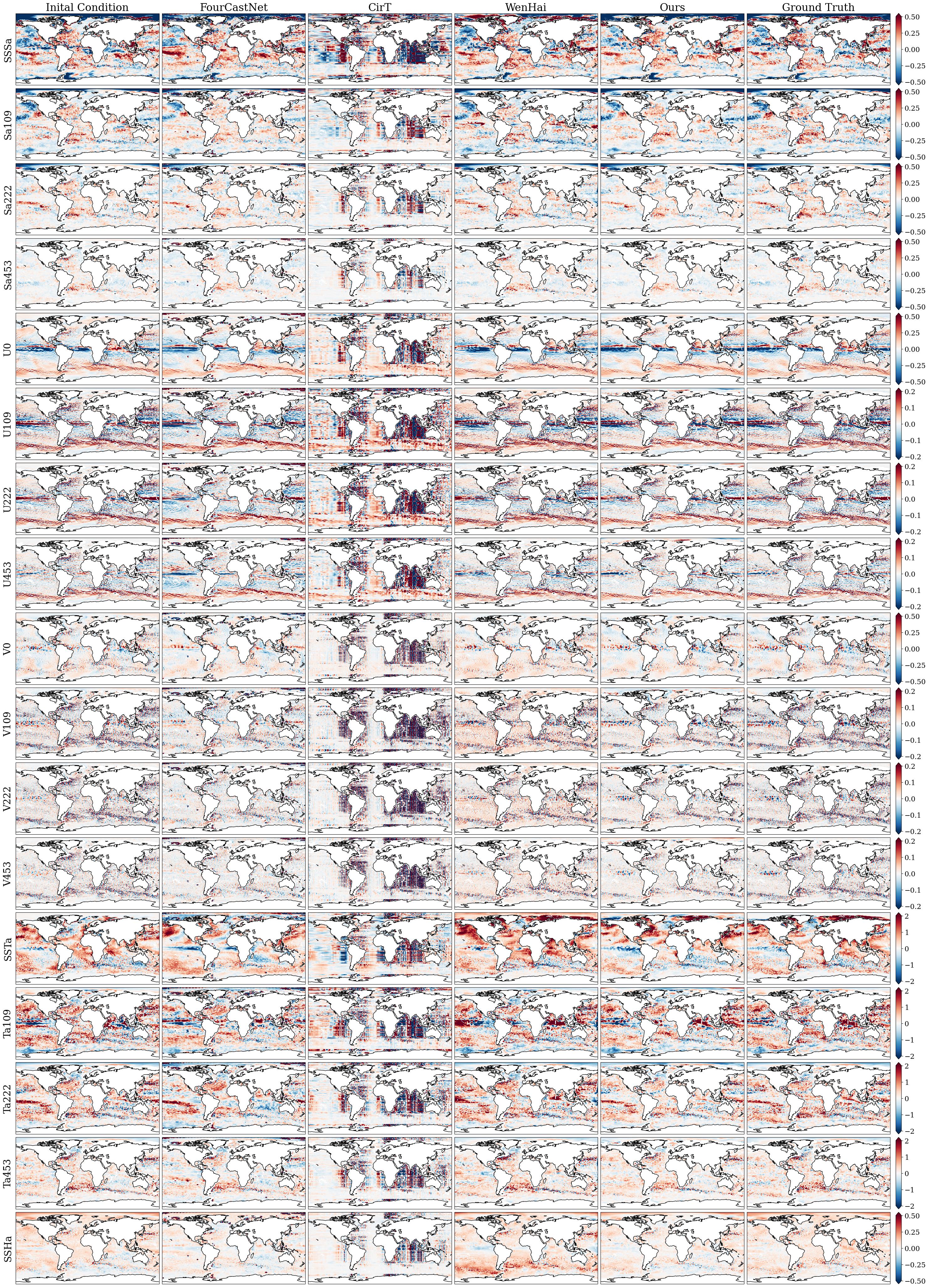}
\caption{60-day simulation results of different models.}
\label{fig_60.0-day}
\end{figure*}

Additional quantitative results (RMSE and ACC across depth levels 109.73 m, 222.48 m, and 453.94 m) are shown in Figure \ref{fig_acc_rmse_appendix}. Due to the relatively slow vertical variations in the ocean, the differences between adjacent layers at similar depths are relatively small. Therefore, selecting representative depth levels allows for a more intuitive assessment of the model's ability to simulate the three-dimensional structure of the ocean. RMSE and ACC represent the average performance for 240 ICs, starting from Jan. 1, 2020, and the interval of each IC is 1 day. Additional visual results are presented in Figure \ref{fig_10.0-day}, Figure \ref{fig_20.0-day}, Figure \ref{fig_30.0-day}, Figure \ref{fig_40.0-day}, Figure \ref{fig_50.0-day}, and Figure \ref{fig_60.0-day}, covering 17 important ocean variables across simulation time from 10 to 60 days, with the initial condition at Jun. 1, 2020. Specifically, for variables S, T, and SSH, where variations are subtle relative to their baseline values, anomalies are presented to highlight these minor but significant changes. Conversely, the original data are directly displayed for U and V. For clarity, examples of variable naming conventions are as follows: SSSa denotes sea surface salinity anomaly, Sa109 denotes salinity anomaly at 109.73 m, Sa222 denotes salinity anomaly at 222.48 m, and Sa453 denotes salinity anomaly at 453.94 m. U0 denotes the sea surface stream zonal velocity, U109 denotes the sea stream zonal velocity at 109.73 m, U222 denotes the sea stream zonal velocity at 222.48 m, and U453 denotes the sea stream zonal velocity at 453.94 m. Collectively, these additional results further demonstrate the efficacy of the proposed NeuralOM.



\end{document}